\documentclass[letterpaper, 10 pt, conference]{ieeeconf}  

\IEEEoverridecommandlockouts                              

\usepackage{algorithm}
\usepackage{algorithmicx}
\usepackage{algpseudocode}
\usepackage{booktabs} 
\usepackage{multirow} 
\usepackage{amssymb,amsmath,comment}

\usepackage{array}        
\usepackage{makecell}     

\usepackage{comment}
\usepackage{listings}
\usepackage{graphicx}
\usepackage{subcaption}
\usepackage{hyperref}

\makeatletter
\def\section{\@startsection{section}{1}{\z@}{1.1ex plus 1ex minus 0.5ex}%
{0.5ex plus .5ex minus 0ex}{\normalfont\normalsize\centering\scshape}}%
\def\subsection{\@startsection{subsection}{2}{\z@}{1.1ex plus 1ex minus 0.5ex}%
{0.5ex plus .3ex minus 0ex}{\normalfont\normalsize\itshape}}%
\makeatother

\title{\LARGE \bf 
VAMP-MR: Vector-Accelerated Motion Planning and Execution for Multi-Robot-Arms
}
\author {
    Philip Huang$^{1}$,
    Chenrui Gao$^{2}$,
    Jiaoyang Li$^{1}$
    \thanks{$^{1}$Philip Huang and Jiaoyang Li is with the Robotics Institute, 
    Carnegie Mellon University, Pittsburgh, PA 15213, USA.
    Corresponding email: {\tt\small philiphuang@cmu.edu}}%
    \thanks{$^{2}$Chenrui Gao is with the University of Michigan, Ann Arbor.}%
} 

\begin{document}

\maketitle

\begin{abstract}
Multi-robot-arm motion planning is a key challenge in deploying multiple manipulators for industrial tasks such as manufacturing. Existing search-based and sampling-based solvers often require significant computation time to produce collision-free, high-quality motions suitable for safe real-world execution. In this work, we introduce a new suite of multi-robot-arm motion planners capable of near real-time motion generation, combining classical planning algorithms with state-of-the-art vectorized collision-checking techniques. Based on CPU SIMD instructions, our new planners accelerate their primary bottleneck, collision checking, and achieve up to two orders of magnitude speedup in both motion planning and execution postprocessing for multi-arm manipulation tasks. 
We also release our implementation to lower the barrier for research and development of multi-robot-arm planning and manipulation problems. Code is available at \url{https://vamp-mr.github.io/vamp-mr}
\end{abstract}

\section{Introduction}
\begin{figure}
    \centering
    \begin{subfigure}[b]{0.88\columnwidth}
        \includegraphics[width=\linewidth]{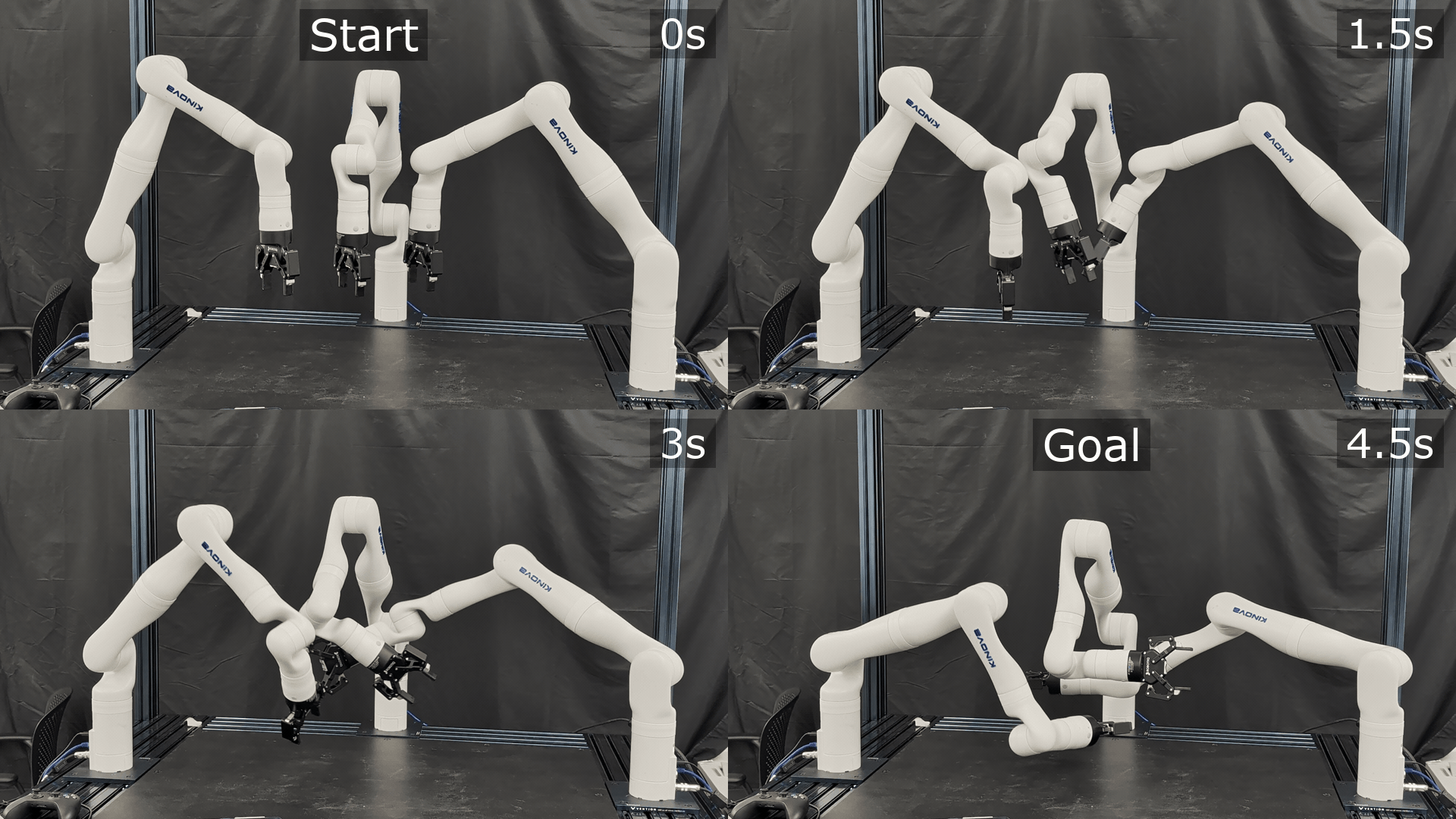}
        \caption{Planning Time: 0.157s, Shortcut Time: 1s}
    \end{subfigure}
    \begin{subfigure}[b]{0.88\columnwidth}
        \includegraphics[width=\linewidth]{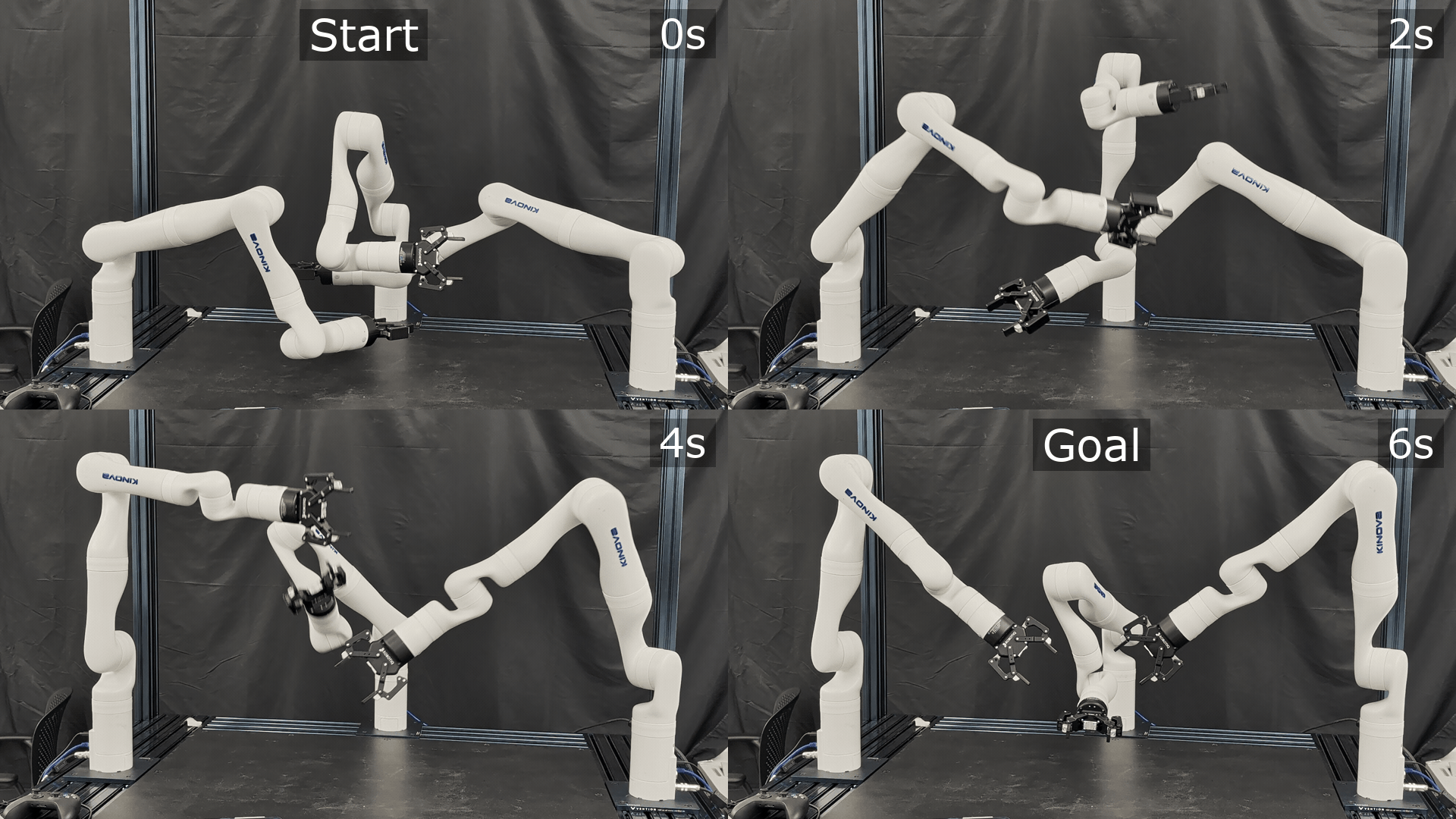}
        \caption{Planning Time: 0.162s, Shortcut Time: 1s}
    \end{subfigure}
    \caption{Examples of multi-robot-arm trajectories planned with our vector-accelerated motion planner. We use composite RRT-Connect~\cite{RRT-Connect} and postprocess the trajectories with 1s of randomized shortcutting~\cite{Huang2025BenchmarkingShortcut}.  }
    \label{fig:teaser}
    \vspace{-0.2cm}
\end{figure}

Robotic systems have the potential to transform industries such as construction and manufacturing by automating hazardous and physically demanding tasks. In theory, deploying multiple robots within a shared workcell (such as Fig. \ref{fig:teaser}) can greatly enhance efficiency and throughput through collaboration and parallelization for tasks such as assembly. In practice, however, coordinating teams of robot arms to operate safely, efficiently, and robustly is highly challenging, and most multi-robot setups still require extensive manual programming that is difficult to adapt when tasks or environments change.

\begin{figure*}[t] 
    \centering
    \begin{subfigure}[b]{0.48\textwidth}
        \centering
        \includegraphics[width=\textwidth]{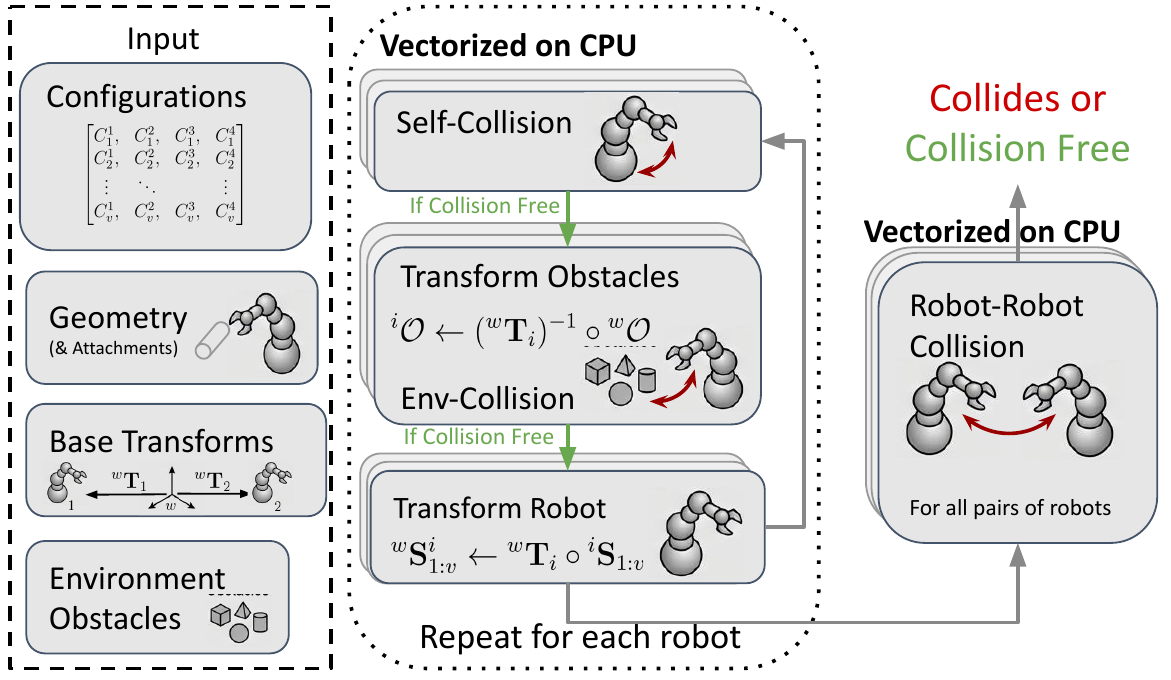}
        \caption{Workflow of our VAMP-MR Procedure}
        \label{fig:vamp-mr}
    \end{subfigure}
    \hfill 
    \begin{subfigure}[b]{0.48\textwidth}
        \centering
        \includegraphics[width=\textwidth]{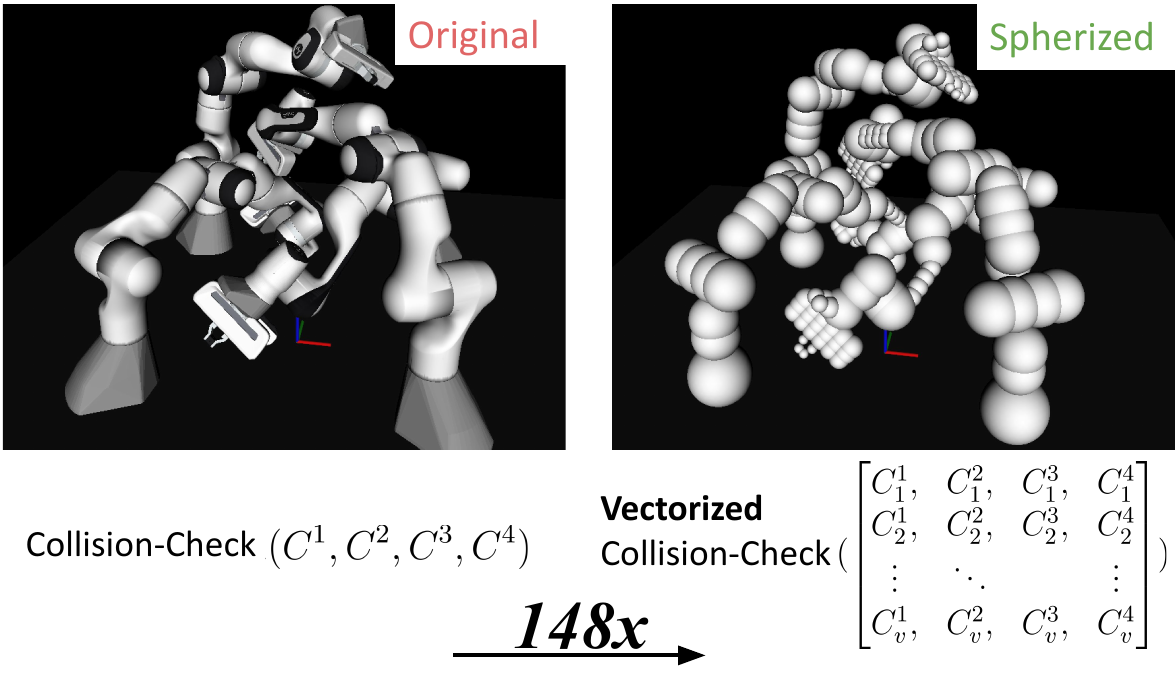}
        \caption{Speedup on Panda Four Arms}
        \label{fig:panda_four}
    \end{subfigure}
    
    \caption{Description of our multi-robot-arm collision checking primitive. Given environment geometries and a batch of multi-robot configurations, VAMP-MR can determine if any collisions exist within the batch up to two orders of magnitude faster than FCL-based methods. The key is to use spherized geometries and leverage SIMD instructions to parallelize over the batch, along with a structured arrangement of self, robot-environment, or robot-robot checks for early collision detection.}
    \label{fig:main_figure}
    \vspace{-4pt}
\end{figure*}

In this work, we target a key computational bottleneck shared by nearly all multi-robot-arm motion planning (M-RAMP) methods—collision checking.
As we discuss in Sec.~\ref{sec:background}, collision checking dominates the runtime of almost every stage of a multi-robot-arm planning and execution pipeline: constructing roadmaps, validating sampled motions, detecting inter-robot collisions, shortcutting trajectories, and coordinating safe execution.
Our approach builds upon Vector-Accelerated Motion Planning (VAMP) \cite{vamp_2024}, a high-performance single-robot-arm motion planner featuring forward kinematics and collision checking vectorized with CPU SIMD (Single Instruction, Multiple Data) instructions, which process a batch of robot configurations in parallel (Sec.~\ref{sec:accel-background}); we extend it to multi-robot settings.
As shown in Fig. \ref{fig:main_figure}, we introduce a general-purpose vectorized collision-checking module for multiple robot arms with flexible input definitions for base transforms, robot count, attachments, obstacles, and configurations.
It serves as a drop-in replacement for collision-checking frameworks such as FCL \cite{pan2012fcl} and Bullet \cite{bullet_2021}, achieving up to 100× speedups in both motion planning and postprocessing without any other algorithmic changes. Our experiments demonstrate that standard planning and postprocessing pipelines can generate the motion for four robot arms in under a second on average, and that existing multi-agent pathfinding (MAPF) algorithms perform well even with minimal adaptation to robot arms.

\section{Background}
\label{sec:background}

\subsection{Problem Definition}
\label{sec:problem}
Multi-Robot-Arm Motion Planning (M-RAMP) is the problem of centrally coordinating the motion of multiple robot arms in tight, cluttered environments.
We are given $N$ robot arms, where robot $i \in \{1, \dots, N\}$ has a fixed base transform ${}^{w}\mathbf{T}_{i} \in SE(3)$ in the world frame and a configuration space $\mathcal{C}^i \subseteq \mathbb{R}^{\text{DoF}^i}$, where $\text{DoF}^i$ is the number of degrees of freedom of robot $i$, and all robots share an environment with static obstacles ${}^{w}\mathcal{O}$.
To model manipulation, each robot may rigidly hold attachments $\mathcal{A}^i$ (e.g., grasped objects or tools) that move with its end effector, and a set of allowed contacts $\mathcal{W}_\text{allow}$ specifies robot-obstacle pairs that are intentionally permitted to touch (e.g., a gripper grasping an object) and are ignored during collision checking.
A \emph{composite configuration} $(C^1, \dots, C^N)$, one configuration per robot, is \emph{collision-free} if every robot, including its attachments, has no self-collision, no collision with obstacles except the allowed contacts in $\mathcal{W}_\text{allow}$, and no collision with other robots.

The objective of M-RAMP is to find a set of collision-free trajectories $\tau = \{\tau^1, \tau^2, \dots \tau^N\}$ from each robot's start configuration $C^i_\text{start}$ to its goal configuration $C^i_\text{goal}$.
Each trajectory $\tau^i$ is represented as a sequence of waypoint configurations $[C^i_1, C^i_2, \dots]$ discretized at a fixed timestep, with linear interpolation between consecutive waypoints.
We measure solution quality by \emph{makespan}, the total execution time until the last robot reaches its goal; for execution on real robots under timing uncertainties, plans are often converted into a temporal plan graph (TPG; Sec.~\ref{sec:manip-background}).

\subsection{Multi-Robot-Arm Motion Planning}
\label{sec:mramp-background}
A straightforward approach to M-RAMP is to treat all robots as a single composite system whose configuration space is the Cartesian product of the individual configuration spaces, and to apply a single-robot planner such as RRT-Connect \cite{RRT-Connect} or graph-of-convex-sets planning \cite{marcucci2022motion}; however, this can be computationally prohibitive as the dimensionality of the composite space grows with the number of robots.
To address scalability, researchers have adapted multi-agent pathfinding (MAPF) techniques \cite{stern2019multi}—originally developed for 2D discrete grid worlds—to the high-dimensional configuration spaces of robot arms.
A prominent example is conflict-based search (CBS) \cite{sharon2015conflict}, a complete and optimal two-level algorithm: the low level plans a path for each agent independently, and the high level resolves collisions between agents by expanding a binary constraint tree, whose child nodes each force one of the two colliding agents to avoid the collision and replan.
Applying MAPF to robot arms requires discretizing each arm's configuration space into a graph, e.g., a probabilistic roadmap (PRM) \cite{PRM}.
CBS-MP \cite{solis2021representation} runs CBS over the individual roadmaps of each arm, and follow-up work accelerates the search with experience reuse \cite{shaoul2024accelerating, shaoul2024gencbs}, retaining completeness and optimality guarantees with respect to the roadmaps.
Alternatively, prioritized planning \cite{Hartmann_rearrange, chen2022cooperativeMRAMP} plans arms sequentially while treating previously planned arms as moving obstacles—often fast in practice but without the guarantees of CBS-style solvers.
Learning-based approaches such as RoboBallet \cite{roboballet-2025} instead train centralized neural networks with reinforcement learning to control synchronized multi-arm motion.

Across all these planning paradigms, collision checking is a dominant computational bottleneck that appears in every part of the pipeline: roadmap construction validates every sampled configuration and candidate edge against self-collisions and environment obstacles; sampling-based planners spend most of their runtime validating sampled configurations and the motions connecting them \cite{Bialkowski2011-jg, vamp_2024}; and CBS-style solvers must detect, and often count, inter-robot collisions between arm trajectories to select collisions \cite{solis2021representation}.
Moreover, because planners rarely return optimal solutions in continuous spaces given finite time, their trajectories are commonly postprocessed by \emph{shortcutting} \cite{choset2005shortcutting, Huang2025BenchmarkingShortcut}, which repeatedly attempts to replace trajectory segments with shorter collision-free connections—a process consisting almost entirely of collision checking.
Our method is complementary to all these works: it accelerates the collision checking shared by sampling- and search-based planners, plus the feature and reward evaluations of learning-based training.

\subsection{Multi-Robot-Arm Manipulation Planning and Execution}
\label{sec:manip-background}
Long-horizon manipulation tasks, such as collaborative assembly \cite{chen2022cooperativeMRAMP, huang2025apexmr}, further magnify the cost of collision checking: in the dual-arm LEGO assembly tasks of APEX-MR \cite{huang2025apexmr}, for example, hundreds of pick, transit, and place motions must be planned sequentially, each in a different \emph{mode} of the environment—with different objects attached to the grippers and different obstacle geometry as placed objects accumulate—so the cost of collision checking increases with longer task horizons.

Executing such plans in the real world must additionally address uncertainties due to controller and sensor delays, since a robot that falls behind its planned timing may collide with another robot that proceeds on schedule.
One solution is the temporal plan graph (TPG) \cite{Honig2019-gr, huang2025apexmr}, a directed acyclic graph in which nodes represent waypoints along each robot's trajectory and directed edges encode precedence constraints: intra-robot edges connect consecutive waypoints of the same robot, while inter-robot edges require one robot to pass a waypoint before another robot may enter a region that would otherwise collide with it.
During execution, a scheduler dispatches a node only after all its predecessors are completed, preserving safety regardless of delays.
APEX-MR \cite{huang2025apexmr} extends the TPG to multi-modal manipulation plans and further reduces the execution makespan with randomized TPG shortcutting.
Here, too, collision checking dominates: identifying inter-robot dependencies during TPG construction requires collision checks between all pairs of trajectory waypoints from different robots—the number of checks scales quadratically with both the number of robots and the trajectory length—and TPG shortcutting must validate every candidate shortcut.

\subsection{Accelerated Motion Planning}
\label{sec:accel-background}
Numerous strategies have been proposed to reduce the burden of collision checking (\textsc{CC}) in motion planning~\cite{Bialkowski2011-jg, Murray2016-xz, vamp_2024}: lazy evaluation delays collision checking until necessary (e.g., LazyPRM~\cite{lazyprm-2000}); experience-based methods exploit past search data~\cite{shaoul2024accelerating}; learning-based approaches train approximate collision detectors~\cite{Das2020-hh} or signed distance fields~\cite{Koptev2023-id}; and neural motion planners such as MPNet~\cite{mpnet-2021} amortize planning cost through large-scale training.

Our work aligns with approaches that directly parallelize collision checking itself on multi-core CPUs \cite{Ichnowski2014-dj}, GPUs \cite{Bialkowski2011-jg}, or robot-specific FPGA circuits \cite{Murray2016-xz}. Offloading to a GPU or FPGA, however, introduces nontrivial communication overhead: a planner running on the CPU must repeatedly query the device for motion feasibility, and the accumulated data-transfer latency can negate much of the speedup.
An alternative is to exploit the data parallelism within a single CPU core: SIMD instructions apply one arithmetic operation simultaneously to a short vector of values packed in a wide register—e.g., 256-bit AVX2 instructions process 8 single-precision floating-point values, referred to as 8 SIMD \emph{lanes}, in one instruction—without any device communication overhead.
We build upon VAMP \cite{vamp_2024}, which batches multiple configurations into the SIMD lanes to evaluate forward kinematics and collision checking for all of them in parallel.

\subsection{Vector-Accelerated Motion Planning (VAMP)}
\label{sec:vamp-background}

Traditional collision checkers such as FCL~\cite{pan2012fcl} and Bullet~\cite{bullet_2021} are primarily designed for single-threaded execution: each collision object is represented by a bounding volume hierarchy (BVH) over its geometric primitives, and checking proceeds in two stages—a broad-phase filter that identifies potentially overlapping bounding volumes, followed by a narrow-phase test (e.g., GJK~\cite{Gilbert1988-oj}) for exact intersections.
While widely adopted, this design is inherently difficult to parallelize: the recursive BVH traversal introduces significant conditional branching and irregular memory access, and the bandwidth to fetch BVH transforms and geometry often becomes limiting.

VAMP~\cite{vamp_2024} overcomes these challenges by fusing forward kinematics (\textsc{FK}), which computes the poses of a robot's links and geometry from a configuration, and \textsc{CC} into a single optimized kernel implemented as a C++ header.
It reorganizes a batch of joint-space configurations—one configuration per SIMD lane—into a struct-of-arrays memory layout for data parallelism.
Robot geometries are approximated by sets of spheres~\cite{Coumar2025-zv}, and obstacles by simple primitives (spheres, cubes, cylinders, or capsules).
\textsc{FK} is unrolled via a custom tracing compiler that computes the positions of these spheres directly, minimizing branching and data dependencies, and self-collision checks are interleaved with \textsc{FK} for early termination.
Computing \textsc{FK} and \textsc{CC} for each sphere is fully parallelized over the batch, and if any configuration collides, the entire batch is rejected.
With batching, VAMP discards traditional broad-phase \textsc{CC} and instead opts for a ``rake" strategy that evaluates a set of uniformly distanced configurations along an edge in one batch, maximizing the likelihood of detecting collisions early for that edge. VAMP's vectorized routines integrate with search- or sampling-based planners for millisecond-level planning.

However, each VAMP kernel is compiled for a single, fixed robot (Sec.~\ref{sec:method}).
We therefore bring these ideas to VAMP-MR, a vectorized multi-robot-arm collision checker that accelerates multiple stages of the pipeline, including motion planning, trajectory shortcutting, and safe execution.


\section{Vectorized Multi-Robot Collision Checking}
\label{sec:method}
\subsection{Method}
Motivated by the pervasive cost of collision checking across the multi-robot planning and execution stack (Sec.~\ref{sec:background}), our core contribution is a vectorized multi-robot collision checking routine that is \emph{efficient}, \emph{flexible}, and \emph{easy to use}.
It is efficient because, like VAMP, it fuses \textsc{FK} and \textsc{CC} over a SIMD batch of configurations with approximate sphere-based robot modeling.
It is flexible because it distinguishes self-, robot-environment, and robot-robot collisions and supports attachments and environment obstacles, matching the diverse collision queries required by the planners in Sec.~\ref{sec:background}.
It is easy to use because collision pairs can be added or disabled and base transforms recalibrated at runtime without kernel recompilation.


These properties do not come for free: a naive adaptation of VAMP to multi-robot planning would merge all DoFs across robots into a single composite system and generate a corresponding monolithic kernel, which cannot separate collision types (e.g., for roadmap generation), count inter-robot collisions between selected robot pairs (e.g., in CBS-style planning and TPG construction), or update ${}^{w}\mathbf{T}_{i}$, $\mathcal{A}^i$, and $\mathcal{W}_\text{allow}$ (Sec.~\ref{sec:problem}) without recompilation.

\begin{algorithm}[t]
\footnotesize
\caption{\textsc{FK\_CC\_Multi}: Vectorized \textsc{FK} and Collision Checking for Multiple Robots}
\label{alg:fk-cc-multi}
\begin{algorithmic}[1]
\Require For each robot $i$: base transform ${}^{w}\mathbf{T}_{i}$, batch of configurations $\{C^i_j\}_{j=1}^{v}$, optional attachments $\mathcal{A}^i$;
environment obstacles ${}^{w}\mathcal{O}$; allowed contacts $\mathcal{W}_{allow}$

\For{each robot $i$}
    \State ${}^{i}\mathbf{S}_{1:v} \leftarrow \textsc{FK}(C^i_{1:v}, \mathcal{A}^i)$ \label{line:fk} \Comment spheres in robot-$i$ frame
    \If{\textsc{SelfCC}$({}^{i}\mathbf{S}_{1:v})$ detects collision} \label{line:selfcc}
        \State \Return \textbf{Invalid}
    \EndIf

    \State ${}^{i}\mathcal{O} \leftarrow ({}^{w}\mathbf{T}_{i})^{-1} \circ {}^{w}\mathcal{O}$ \label{line:transform-obs} \Comment obstacles in robot-$i$ frame
    \If{\textsc{EnvCC}$({}^{i}\mathbf{S}_{1:v}, {}^{i}\mathcal{O}, \mathcal{W}_{allow})$ detects collision} \label{line:envcc}
        \State \Return \textbf{Invalid}
    \EndIf

    \State ${}^{w}\mathbf{S}^{i}_{1:v} \leftarrow {}^{w}\mathbf{T}_{i} \circ {}^{i}\mathbf{S}_{1:v}$ \label{line:world} \Comment spheres in world frame
\EndFor

\For{each robot pair $(i,k),\ i<k$} \label{line:pair}
    \If{\textsc{InterCC}$({}^{w}\mathbf{S}^{i}_{1:v}, {}^{w}\mathbf{S}^{k}_{1:v})$ detects collision} \label{line:intercc}
        \State \Return \textbf{Invalid}
    \EndIf
\EndFor

\State \Return \textbf{Valid} \label{line:valid} \Comment all $v$ composite configurations in the batch are collision-free
\end{algorithmic}
\end{algorithm}

To support these capabilities, we design a new routine, \textsc{FK\_CC\_Multi}, outlined in Alg.~\ref{alg:fk-cc-multi} and illustrated in Fig.~\ref{fig:vamp-mr}.
Its input is a batch of $v$ composite configurations $\{(C^1_j, \dots, C^N_j)\}_{j=1}^{v}$, where $v$ is the SIMD batch size; we use the subscript $1{:}v$ to denote a batch, e.g., $C^i_{1:v} = [C^i_1, \dots, C^i_v]$ is the batch of $v$ configurations of robot $i$.
\textsc{FK\_CC\_Multi} computes forward kinematics and checks collisions in one call over the whole batch, and its returned validity flag is \textbf{Valid} if and only if all $v$ composite configurations in the batch are collision-free.
Each robot first passes through the single-robot vectorized \textsc{FK} routine, which computes the robot's collision spheres in its base frame, followed by a self-collision check (Lines \ref{line:fk}-\ref{line:selfcc}). If a collision is detected, the whole batch is immediately rejected.
Next, environment obstacles ${}^w\mathcal{O}$ are transformed into the robot's base frame and checked against the robot's spheres, ignoring any allowed contacts in $\mathcal{W}_\text{allow}$ (Lines \ref{line:transform-obs}-\ref{line:envcc}).
In practice, we cache the transformed obstacles in each robot's base frame since base transforms and obstacles often remain fixed during planning.
Then, each robot's spheres are transformed to the world frame based on its base transform ${}^{w}\mathbf{T}_{i}$ (Line \ref{line:world}).
Finally, the sphere representations of all robots are compared pairwise to detect robot-robot collisions (Lines \ref{line:pair}-\ref{line:intercc}).

\begin{figure}[t!]
\centering
\includegraphics[width=\linewidth]{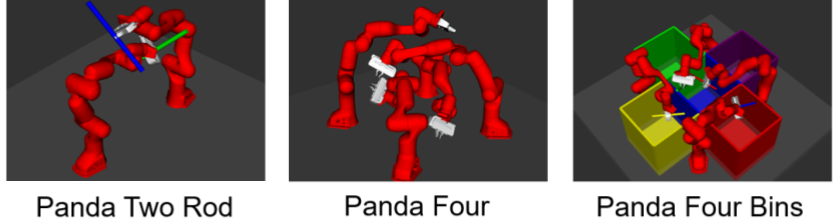}
\caption{Multi-robot motion planning environments.}
\label{fig:robot_envs}

\end{figure}

\subsection{Evaluation of Collision Checking and Motion Validation}
To evaluate the runtime improvement of our new collision checking method, we test on three challenging multi-robot-arm environments—Panda Two Rod, Panda Four, and Panda Four Bins (see Fig. \ref{fig:robot_envs})—introduced in \cite{Huang2025BenchmarkingShortcut}.
Each Panda arm has 7 DoF, and the 59-sphere approximation in Fig. \ref{fig:panda_four} is generated with the tool in \cite{Coumar2025-zv}.
Panda Two Rod adds a rod attachment to each arm, and Panda Four Bins adds bins as environment obstacles that each arm must reach into.

Table~\ref{tab:checking_speed_narrow} reports the speedup of our vectorized collision-checking primitives over 10,000 random samples for both (1) collision checking of a single composite configuration (i.e., $v = 1$) and (2) motion validation between a pair of composite configurations, where the interpolated configurations along the motion fill the SIMD batch ($v = 8$).
For motion validation, the collision checking resolution, defined as the $L_1$ distance between two consecutive interpolated configurations along the motion, is set to 0.1 radian.
We compare against FCL~\cite{pan2012fcl}, the default collision checker of the widely used MoveIt framework~\cite{coleman2014reducing} and of prior multi-arm planning systems~\cite{huang2025apexmr, solis2021representation}; Bullet~\cite{bullet_2021} follows a similar BVH-based, single-query design, so we expect similar trends.
To ensure consistent modeling, we use the same simplified spherized robot geometry representation in FCL.
All experiments run on an AMD 7840HS laptop CPU, with C++ compiled with GCC 9 and 256-bit AVX2 instructions (-march=native, -mavx2, -O3), giving a SIMD batch size of 8 with single-precision floats.
We use AVX2 rather than the wider AVX-512 because, as observed by the authors of VAMP~\cite{vamp_2024}, 512-bit instructions can lower the CPU clock frequency due to power and thermal limits, making AVX2 the better balance.

Our method achieves 11-27x speedup for single-configuration collision checking, and up to 148x speedup for motion validation between two random configurations.
For motion validation, the runtime speedup sometimes exceeds the single-check speedup by more than 8 times (i.e., the number of SIMD lanes), which we attribute to the combined effect of vectorization, good cache utilization, and the ``rake"-style scan over discretized configurations.
We also observe less runtime variance across environments than FCL, which invalidates the mostly colliding random motions in obstacle-rich Panda Four Bins much faster than elsewhere.

\section{Applications of Vector-Accelerated Collision Checking}
\subsection{Vectorized Multi-Robot-Arm Planning}
We integrate our vectorized collision-checking framework into two multi-robot-arm motion planners. Our first planner is composite RRT-Connect~\cite{RRT-Connect}, which combines the DoFs from all robots and treats them as a single robot for the RRT-Connect algorithm. Our \textsc{FK\_CC\_Multi} routine accelerates the validation of randomly sampled configurations and motions during tree expansion.

\begin{table}[t!]
\caption{Average collision checking runtime and speedup over 10,000 random samples. Our method is compared against FCL with the same approximated spherized geometries to ensure consistency.}
\label{tab:checking_speed_narrow}
\centering
\footnotesize
\setlength{\tabcolsep}{6pt} 
\begin{tabular}{l c r r r}
\toprule
\textbf{Check} & \textbf{Environment} & \textbf{FCL (\textmu s)} & \textbf{Ours (\textmu s)} & \textbf{Speedup} \\
\midrule
\multirow{3}{*}{Single} 
 & Panda Two Rod  & 100.99  & {8.60} & \textbf{11.7x} \\
 & Panda Four   & 206.53  & {15.01} & \textbf{13.7x} \\
 & Panda Four Bins & 382.63  & {13.70} & \textbf{27.9x} \\
\midrule
\multirow{3}{*}{Motion} 
 & Panda Two Rod  & 3936.13 & {36.03} & \textbf{109.2x} \\
 & Panda Four      & 4836.50 & {32.61} & \textbf{148.3x} \\
 & Panda Four Bins & 930.03  & {14.23} & \textbf{65.4x} \\
\bottomrule
\end{tabular}
\end{table}

\begin{figure*}[h]
    \centering
    \includegraphics[width=0.95\linewidth]{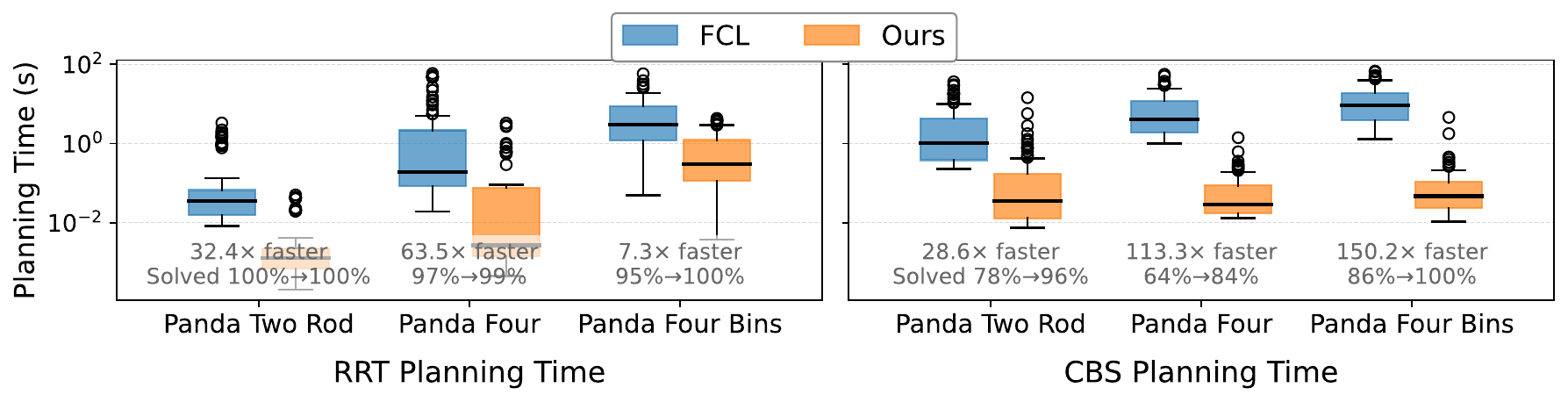}
    \caption{Planning time comparison between FCL-based and our VAMP-based motion planners. Boxplots show the distribution of planning times over solved instances. The speedup reported for each environment is the ratio between the median planning times of the FCL-based and VAMP-based planners over their respective solved instances, and is up to 150x.}
    \label{fig:planning_time}
    
\end{figure*}


Our second planner is based on CBS-MP~\cite{solis2021representation}, which applies CBS to robot arms, each planning on its own PRM and resolving inter-robot collisions through the high-level constraint tree with avoidance constraints (Sec.~\ref{sec:mramp-background}).
However, the original avoidance constraint used in CBS-MP is incomplete, as discussed in \cite{shaoul2024accelerating}.
To ensure theoretical completeness of CBS on the roadmaps, we adopt an asymmetric constraint scheme inspired by \cite{li2019multi}.
Specifically, when a collision occurs at time $t$ between robot $i$ and robot $j$, we create two mutually exclusive constraints during high-level CBS expansion: one forbidding robot $i$ from occupying any volume of robot $j$'s volume at time $t$, and the other forbidding robot $j$ from taking the same configuration at that time.
Our SIMD-accelerated \textsc{FK\_CC\_Multi} routine reduces the computational overhead of all three collision-related components of CBS-MP: (1) roadmap construction, which requires only self- and robot-environment collision checks when validating each arm's sampled configurations and edges; (2) conflict detection, which checks robot-robot collisions between the trajectories of different arms to identify collisions to resolve; and (3) constraint evaluation in the low-level search, which checks a robot's motion against the constrained volumes of other robots.

\begin{table}[t]
\centering
\caption{Cache behavior comparison between FCL-based and our VAMP-based motion planners. We report the cache miss rate and cache misses per 1,000 instructions (MPKI).}
\label{tab:cache_mpki}
\footnotesize
\setlength{\tabcolsep}{4pt}

\begin{tabular}{c c cc cc}
\toprule
\multirow{2}{*}{\textbf{Environment}} & \multirow{2}{*}{\textbf{Planner}} 
& \multicolumn{2}{c}{\textbf{Cache Miss (\%)}} 
& \multicolumn{2}{c}{\textbf{MPKI}} \\
\cmidrule(lr){3-4} \cmidrule(lr){5-6}
 & & FCL & Ours & FCL & Ours \\
\midrule

\multirow{2}{*}{\makecell{Panda Two Rod}}
 & CBS & 4.45 & \textbf{0.79} & 0.83 & \textbf{0.38} \\
 & RRT & \textbf{4.11} & 7.49 & \textbf{0.85} & 2.08 \\
 \midrule
 
\multirow{2}{*}{Panda Four}
 & CBS & 5.49 & \textbf{1.86} & 0.83 & \textbf{0.65} \\
 & RRT & 7.92 & \textbf{2.43} & 1.49 & \textbf{0.55} \\
\midrule

\multirow{2}{*}{\makecell{Panda Four Bins}}
 & CBS & 8.94 & \textbf{2.13} & 1.50 & \textbf{0.83} \\
 & RRT & 13.81 & \textbf{1.72} & 2.62 & \textbf{0.51} \\

\bottomrule
\end{tabular}
\end{table}

\textbf{Results}\quad
To evaluate how VAMP-MR can improve planning time and analyze its algorithmic implications, we create 110, 132, and 132 unique pairs of start and goal poses as different planning instances in Panda Two Rod, Panda Four, and Panda Four Bins, as in \cite{Huang2025BenchmarkingShortcut}.
We evaluate both the composite RRT-Connect and CBS-MP motion planner (abbreviated as RRT and CBS in figures and tables).
We sample 5,000 nodes for each robot's PRM before planning in CBS-MP and run both planners with a one-minute timeout.
Roadmap construction, which also benefits from vectorization, is performed once per environment and excluded from the reported planning time; its speedup is not evaluated.

\begin{figure*}[t]
    \centering
    \includegraphics[width=0.95\linewidth]{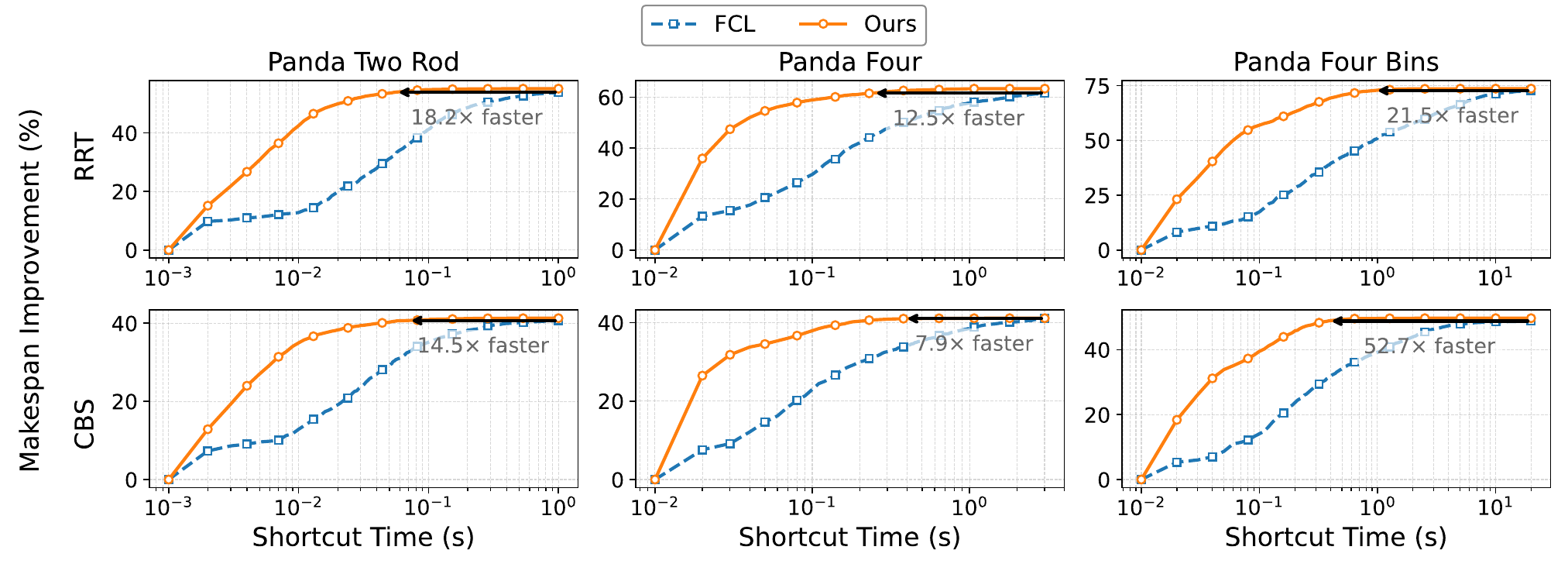}
    \caption{Average makespan improvement over time of multi-robot shortcutting across three environments, using DTS~\cite{Huang2025BenchmarkingShortcut}. The performance-runtime curves are averaged across all instances in each environment. Our method achieves the same level of makespan improvement within one second and is up to 50 times faster than FCL.}
    \label{fig:shortcut}
    
\end{figure*}

Fig. \ref{fig:planning_time} shows the planning time before and after vector acceleration for all three environments. 
Our results show that vectorization provides at least an order-of-magnitude speedup for both planners, with even greater improvement for CBS-MP.
Table~\ref{tab:cache_mpki} compares the cache behavior of FCL and our VAMP collision checker across CBS-MP and RRT-Connect. Except for RRT-Connect in Panda Two Rod, VAMP-MR achieves substantially lower cache miss rates, indicating improved memory locality during frequent collision queries.

For RRT-Connect, the runtime acceleration correlates closely with the motion validation speedup reported in Table \ref{tab:checking_speed_narrow}, as motion validation dominates the computational cost.
However, CBS-MP exhibits the largest performance gain in Panda Four Bins, reaching a 100\% success rate, whereas some instances in Panda Four remain unsolved due to the coordination requirements of the tasks: in Panda Four, several target poses (e.g., the one in Fig. \ref{fig:panda_four}) require precise sequencing between robot arms and thus thousands of high-level CBS expansions, whereas Panda Four Bins requires little coordination as long as each robot can navigate around obstacles independently.
Thus, our vectorized CBS-MP outperforms vectorized RRT-Connect in Panda Four Bins, as it can quickly deconflict agents and find collision-free solutions.
We also noticed that vectorization significantly increases the number of constraint tree (CT) nodes that CBS-MP can expand within the time limit: our vectorized CBS-MP can solve an instance with 5,303 expanded CT nodes on Panda Four, compared to a maximum of 53 expanded CT nodes among solved Panda Four instances with FCL under the same one-minute time limit.

These results demonstrate that our vectorized framework substantially alleviates the primary bottleneck in motion planning—motion validation and collision detection—enabling CBS-based methods to scale to more complex setups.
Nevertheless, even after acceleration, collision checking remains the largest cost, accounting for 90.2\% of planning time for RRT-Connect and 64.6\% for CBS-MP in Panda Four.
Since this remaining cost scales with the number of search expansions, MAPF techniques that guide the search to expand fewer nodes—and hence perform fewer collision checks—such as symmetry reasoning \cite{Li2021-ab}, enhanced bounded-suboptimal search~\cite{li2021eecbs}, and fast suboptimal planners~\cite{li2022mapf}, are promising complements to further reduce the planning time of multi-robot-arm systems.

\subsection{Vectorized Multi-Robot Shortcutting}
Shortcutting (Sec.~\ref{sec:mramp-background}) is a widely used postprocessing technique for improving the smoothness and optimality of planned trajectories in multi-robot-arm motion planning~\cite{shaoul2024accelerating, Hartmann_rearrange}: given finite time, sampling-based planners are only probabilistically complete and asymptotically optimal, and CBS is only resolution-complete and resolution-optimal given the roadmap, so solution quality can usually be improved further with postprocessing.

We define a shortcut as a linear interpolation between two configurations whose endpoints are randomly sampled along the current trajectory; depending on the shortcutting strategy described below, the two configurations may be composite configurations of all robots or configurations of a single robot.
After sampling a shortcut, we check whether replacing the original segment with the shortcut introduces any self-collisions or collisions with other robots or obstacles. If the shortcut is collision-free, the trajectory is updated, and this process repeats until a time limit is reached. Since collision checking dominates the cost of shortcutting, it is highly compatible with our vectorized routine.


We adopt the Dynamic Thompson Shortcutting (DTS) algorithm from~\cite{Huang2025BenchmarkingShortcut}, an anytime shortcutting framework that adaptively combines three complementary strategies: a composite shortcut is sampled in the composite configuration space and jointly modifies all robot trajectories; a prioritized shortcut modifies a single robot's trajectory and shifts the waypoints after the shortcut to earlier timesteps, shortening that robot's trajectory in time; and a path shortcut also modifies a single robot's trajectory but preserves the original timing. DTS formulates the strategy selection as a multi-armed bandit problem and uses Thompson sampling to balance rapid early improvement against final trajectory quality.

\textbf{Results}\quad We evaluate the performance improvement of DTS, as shown in Fig. \ref{fig:shortcut}, using initial trajectories generated by RRT-Connect and CBS-MP.
Without any other modifications, integrating a vectorized collision checking routine reduces the time required to converge to final shortcut trajectories by up to 50x, with substantial improvement within 0.1s and full convergence within 1s.

\subsection{Vectorized Safe Execution Framework}
Safe real-world execution of multi-robot manipulation plans, such as object rearrangement or collaborative assembly, must account for uncertainties such as kinematic inaccuracies and unpredictable physical interactions; many tasks also use closed-loop control or learned policies whose execution times cannot be predicted in advance.

We build upon the multi-modal Temporal Plan Graph (TPG) framework of APEX-MR~\cite{huang2025apexmr}, introduced in Sec.~\ref{sec:manip-background}, which systematically postprocesses a given multi-robot task and motion plan to enable safe, asynchronous execution under such uncertainties.
A multi-modal TPG $G = (V, E)$ extends the TPG in Sec.~\ref{sec:manip-background} to capture kinematic transitions, changes in the environment due to objects being moved, and inter-robot task and motion dependencies.
Each node $v^i_n$ is either a pose node that corresponds to a target configuration $C^i_n$ for robot $i$ or a skill node that corresponds to a manipulation skill, i.e., a set of object-centric motions executed by a feedback controller; we assume each skill has a reference robot trajectory for the purpose of TPG computation.
Each edge $v^i_n \rightarrow v^{i'}_{n'}$ encodes a precedence constraint: edges between nodes of the same robot connect every consecutive node, and inter-robot edges are added for task dependencies or motion dependencies to prevent executing two spatially colliding nodes simultaneously.

\begin{figure}[t]
    \centering
    \includegraphics[width=0.9\linewidth]{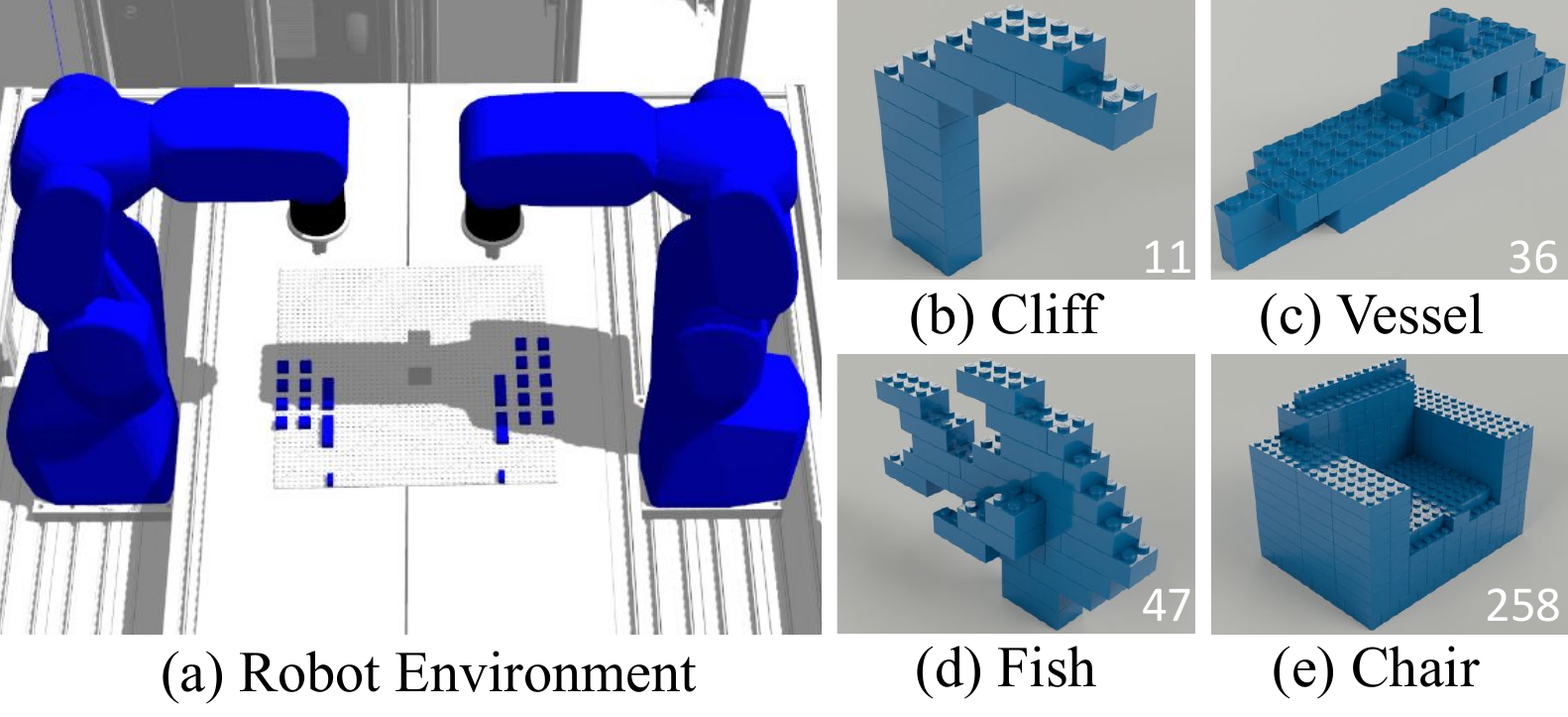}
    \caption{Dual-arm LEGO assembly environment in (a) and assembly tasks in (b)-(e). Numbers in (b)-(e) indicate the number of assembly steps (LEGO bricks).}
    \label{fig:lego_envs}
    \vspace{-0.2cm}
    
\end{figure}
Constructing a TPG from a sequential or synchronous task and motion plan requires finding all inter-robot motion dependencies with quadratically many pairwise collision checks (Sec.~\ref{sec:manip-background}).
Because consecutive pose nodes are densely discretized from the trajectories at the collision checking resolution, checking pairs of node configurations from two robots also covers the motions between them.
While Huang et al. \cite{huang2025apexmr} used CPU multithreading to parallelize this computation, our vectorized routine is even better suited to accelerate TPG construction with less communication overhead.
However, the original TPG formulation requires identifying all colliding pairs of configurations, whereas our vectorized collision checker \textsc{FK\_CC\_Multi} terminates early if any configuration in the batch collides. 
To effectively leverage SIMD parallelism, we propose a modified grouping strategy.

For a group of $k$ consecutive pose nodes $[v^i_n, \dots v^i_{n+k-1}]$ of the TPG, we merge them to a larger transit node $\mathbf{v}^i_n$ that represents a short path segment $[C^i_n, \dots C^i_{n+k-1}]$. We then perform collision checks between every pair of transit nodes from different robots. 
If any part of the path segment of a transit node $\mathbf{v}^i_n$ collides with that of another transit node $\mathbf{v}^{i'}_{n'}$, we insert an inter-robot edge $\mathbf{v}^i_n \rightarrow \mathbf{v}^{i'}_{n'}$ from the earlier node to the later node ($n < n'$) that prevents collision in execution. 
In this formulation, we only need to determine whether any part of the two path segments of two transit nodes $\mathbf{v}^i_n$ and $\mathbf{v}^{i'}_{n'}$ collide, without identifying which individual configurations collide, so the early-terminating batched checks of \textsc{FK\_CC\_Multi} can be applied directly.
This grouping makes the precedence constraints coarser than in the original TPG formulation: a robot must now wait for another robot to complete an entire transit node ($k$ pose nodes) rather than a single pose node, which can introduce additional waiting during execution and thereby increase the makespan. Empirically, however, we find this makespan difference to be negligible.
As a result, our approach achieves much faster TPG construction for complex assembly tasks.

\begin{table}[t!]
\centering
\caption{Runtime and makespan of dual-arm LEGO assembly plans generated following the procedure in APEX-MR~\cite{huang2025apexmr}, with and without vector acceleration. Results are averaged over 4 random seeds. Note that FCL uses 16 CPU threads to parallelize TPG construction, whereas our collision checker is single-threaded.   }
\label{tab:apex-time}
\footnotesize
\begin{tabular}{@{}lrrrr@{}}
\toprule
\textbf{Metric} & \textbf{Cliff} & \textbf{Vessel} & \textbf{Fish} & \textbf{Chair} \\
\midrule
TPG Shortcut Time (s) & 1.0 & 1.0 & 1.0 & 5.0 \\
\midrule
\multicolumn{5}{l}{\textit{FCL-Based}} \\
\quad Task Assignment (s) & 0.60 & 4.90 & 5.81 & 13.0 \\
\quad Motion Planning (s) & 0.63 & 0.76 & 4.91 & 9.98 \\
\quad TPG Construction (s) & 9.88 & 25.6 & 66.0 & 817.2 \\
\quad \textbf{Total Time (s)} & \textbf{12.1} & \textbf{32.2} & \textbf{71.5} & \textbf{845.2} \\
\quad \textbf{Final Makespan (s)} & \textbf{185} & \textbf{397} & \textbf{549} & \textbf{2180} \\
\midrule
\multicolumn{5}{l}{\textit{Ours}} \\
\quad Task Assignment (s) & 0.55 & 4.68 & 5.45 & 10.2 \\
\quad Motion Planning (s) & 0.42 & 0.15 & 1.96 & 0.88 \\
\quad TPG Construction (s) & 1.42  & 2.55 & 3.11 & 43.22 \\
\quad \textbf{Total Time (s)} & \textbf{3.39} & \textbf{8.38} & \textbf{11.8} & \textbf{59.3} \\
\quad \textbf{Final Makespan (s)} & \textbf{159} & \textbf{340} & \textbf{489} & \textbf{2146} \\
\midrule
\# of Bricks & 11 & 36 & 29 & 258 \\
\bottomrule
\end{tabular}
\end{table}

\textbf{Results}\quad We evaluate four long-horizon dual-arm LEGO assembly tasks (see Fig. \ref{fig:lego_envs}) from APEX-MR~\cite{huang2025apexmr}, which involve up to 258 assembled parts. For each task, we measure the runtime across all stages of the pipeline, including task assignment, motion planning, and TPG construction.
We follow the experimental setup of \cite{huang2025apexmr}: given a sequential assembly plan, an integer-linear program assigns robot, grasping, and support pose targets for each assembly step; the motion of each transit task and manipulation skill is planned sequentially with single-agent RRT-Connect and smoothed by a vectorized single-agent shortcutter for 0.1s \cite{choset2005shortcutting}; and the resulting sequential task and motion plan is converted into an asynchronous TPG with TPG shortcutting as in \cite{huang2025apexmr}.

Unlike the original APEX-MR implementation, which plans with the MoveIt framework~\cite{coleman2014reducing}, we implement the RRT-Connect algorithm directly for sequential motion planning for both the FCL-based baseline and our vector-accelerated planner.
The FCL baseline uses the original robot meshes for collision checking, which we found faster than the spherized geometry; our RRT-Connect implementation is also significantly faster than reported in \cite{huang2025apexmr}.

Table \ref{tab:apex-time} summarizes the runtime of each planning step and the final makespan after shortcutting.
Motion planning is accelerated by 1.25-11.3x and TPG construction by 6.9-21x.
In particular, the FCL-based TPG construction uses 16 CPU threads, whereas our vectorized TPG construction uses a single thread—multi-core parallelization is largely unnecessary with vectorization due to its communication overhead.
Our accelerated planner also produces higher-quality solutions after TPG shortcutting: with the same one-second time limit, vectorization achieves more than 10\% lower makespan for cliff, vessel, and fish, while the improvement for chair is minimal due to the large overhead of modifying obstacle positions when evaluating random shortcuts.
Finally, task assignment improves little, as our acceleration affects only the verification of collision-free grasp and support poses; the integer-linear program computing optimal assignments can itself be a significant bottleneck in the overall pipeline.

\section{Real Robot Deployment}
We deploy our vector-accelerated multi-robot-arm motion planner and anytime shortcutter on three Kinova arms in the real world, as shown in Fig.~\ref{fig:teaser}, consecutively planning and executing motions through 11 complex, highly entangled configurations. We use the composite RRT-Connect planner to generate an initial trajectory, post-process it with our anytime shortcutter for 1\,s, and time-parameterize it with TOPPRA~\cite{Pham2018-be} in the composite space to satisfy the arms' acceleration limits. Since the controller delay is minimal in the absence of contact or manipulation policies, the robots execute the planned trajectories synchronously without requiring a TPG.
Across 11 planning calls, the median planning time is 0.16\,s with a sample standard deviation of 0.70\,s; the large standard deviation is due to two difficult instances that required 2.36\,s and 1.06\,s to solve.
The entire demonstration takes about 85\,s, of which all planning combined accounts for only 4.7\,s ($\sim$6\%), with the executed motions averaging about 7\,s each (see the supplementary video).

\section{Conclusion}
We present VAMP-MR, a CPU-accelerated collision checking framework tightly integrated with multi-robot-arm motion planning algorithms.
We deliver 10 to 100x speedup across diverse planning and shortcutting benchmarks without requiring any algorithmic changes, and further accelerate asynchronous planning and execution for long-horizon, dual-arm LEGO assembly tasks.
We believe this work lowers the barrier to developing efficient multi-robot-arm planning algorithms; future directions include integrating advanced MAPF techniques into manipulation and exploring real-time, online planning for long-horizon tasks.


\section*{Acknowledgments}
This work was partially supported by the National Science Foundation under grant numbers \#$2328671$ and \#$2441629$. We thank Zachary Kingston and Wil Thomason for open-sourcing VAMP and for helpful discussions.


\bibliographystyle{ieeetr}
{\footnotesize
\bibliography{ref}
}



\end{document}